\documentclass[10pt,twocolumn,letterpaper]{article}

\usepackage{wacv}
\usepackage{times}
\usepackage{epsfig}
\usepackage{graphicx}
\usepackage{amsmath}
\usepackage{amssymb}
\usepackage{booktabs}
\usepackage[numbers,sort,compress]{natbib}

\def\wacvPaperID{1841} 

\wacvalgorithmstrack   

\wacvfinalcopy 

\ifwacvfinal
\usepackage[breaklinks=true,bookmarks=false]{hyperref}
\else
\usepackage[pagebackref=true,breaklinks=true,colorlinks,bookmarks=false]{hyperref}
\fi

\begin{document}

\title{SoMoFormer: Multi-Person Pose Forecasting with Transformers}

\author{Edward Vendrow\textsuperscript{1}, Satyajit Kumar\textsuperscript{1}, Ehsan Adeli\textsuperscript{1}, and Hamid Rezatofighi\textsuperscript{2}\\
\textsuperscript{1}Stanford University, \textsuperscript{2}Monash University\\
{\tt\small \{evendrow, eadeli\}@stanford.edu, ammesatyajit@gmail.com, hamid.rezatofighi@monash.edu }
}

\maketitle
\thispagestyle{empty}

\begin{abstract}
   Human pose forecasting is a challenging problem involving complex human body motion and posture dynamics. In cases that there are multiple people in the environment, one's motion may also be influenced by the motion and dynamic movements of others. Although there are several previous works targeting the problem of multi-person dynamic pose forecasting, they often model the entire pose sequence as time series (ignoring the underlying relationship between joints) or only output the future pose sequence of one person at a time. In this paper, we present a new method, called Social Motion Transformer (SoMoFormer), for multi-person 3D pose forecasting. Our transformer architecture uniquely models human motion input as a joint sequence rather than a time sequence, allowing us to perform attention over joints while predicting an entire future motion sequence for each joint in parallel. We show that with this problem reformulation, SoMoFormer naturally extends to multi-person scenes by using the joints of all people in a scene as input queries. Using learned embeddings to denote the type of joint, person identity, and global position, our model learns the relationships between joints and between people, attending more strongly to joints from the same or nearby people. SoMoFormer outperforms state-of-the-art methods for long-term motion prediction on the SoMoF benchmark as well as the CMU-Mocap and MuPoTS-3D datasets. Code is available at \url{blinded}. 
\end{abstract}

\section{Introduction}

Forecasting human motion is a challenging problem with applications in robotics \cite{Koppula2013AnticipatingHA}, computer graphics \cite{Levine2012ContinuousCC}, and tracking in the context of autonomous vehicles \cite{Gong2011MultihypothesisMP}. Any system requiring understanding or navigation of multi-person environments requires reasoning about how people will move to make reliable operational decisions.

To date, most of the human skeleton \textit{pose} forecasting methods produce predictions for one person at a time \cite{Chiu2019ActionAgnosticHP,Mao2019LearningTD,Mao2020HistoryRI,Mao2021MultilevelMA,MartinezGonzalez2021PoseT,wang2021multi,Wang2021SimpleBF}, ignoring the global translation and social context of other people in a scene. Simultaneous research into \textit{trajectory} forecasting \cite{Kosaraju2019SocialBiGATMT,Giuliari2021TransformerNF,Sadeghian2019SoPhieAA,Li2020EndtoendCP,Yu2020SpatioTemporalGT,Liu2021MultimodalMP} models an entity as a single 2D point on a ground plane, which provides useful global translation prediction, but is insufficient for tasks requiring detailed body pose information. Recent methods \cite{Adeli2020SociallyAC, Adeli2021TRiPODHT} attempt to model social interactions through pooling and graph-based methods to capture social dependencies for multi-person forecasting, but fall short on long time-horizons with recurrent architectures that quickly accumulate errors. Subsequent methods built on the success of attention-based models for long-term motion prediction \cite{MartinezGonzalez2021PoseT,Giuliari2021TransformerNF}, with \cite{MartinezGonzalez2021PoseT} circumventing the sequential nature of time-series pose forecasting with a Transformer architecture capable of predicting an entire single-person pose trajectory in parallel. Yet, since their model used a fixed input size with a time-series input, generalization to the multi-person setting is difficult.

\begin{figure}[t]
    \centering
    \includegraphics[width=\linewidth]{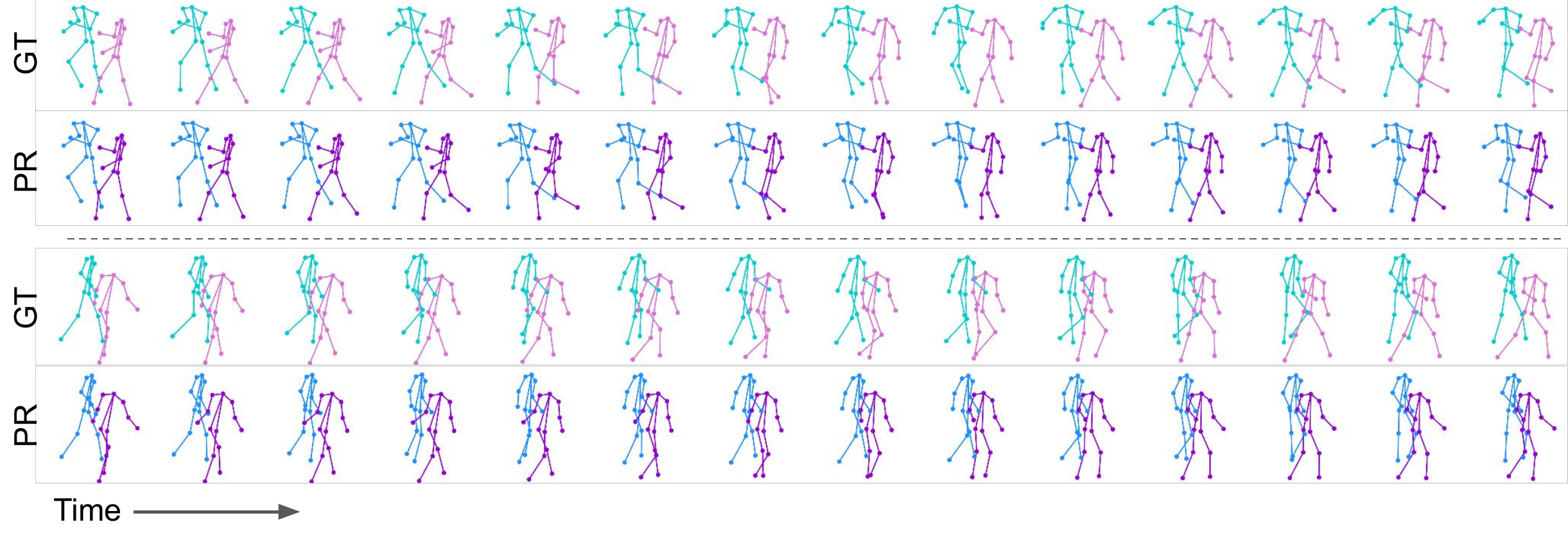}
    \caption{Visualization of SoMoFormer predictions on the 3DPW dataset for two sequences, with the ground-truth future sequence (GT) on top and predicted sequence (PR) below. SoMoFormer predicts realistic human motion, closely match the ground-truth even in a multi-person scenario.}
    \label{fig:visual}
\end{figure}

\begin{figure*}[t]
    \centering
    \includegraphics[width=\linewidth]{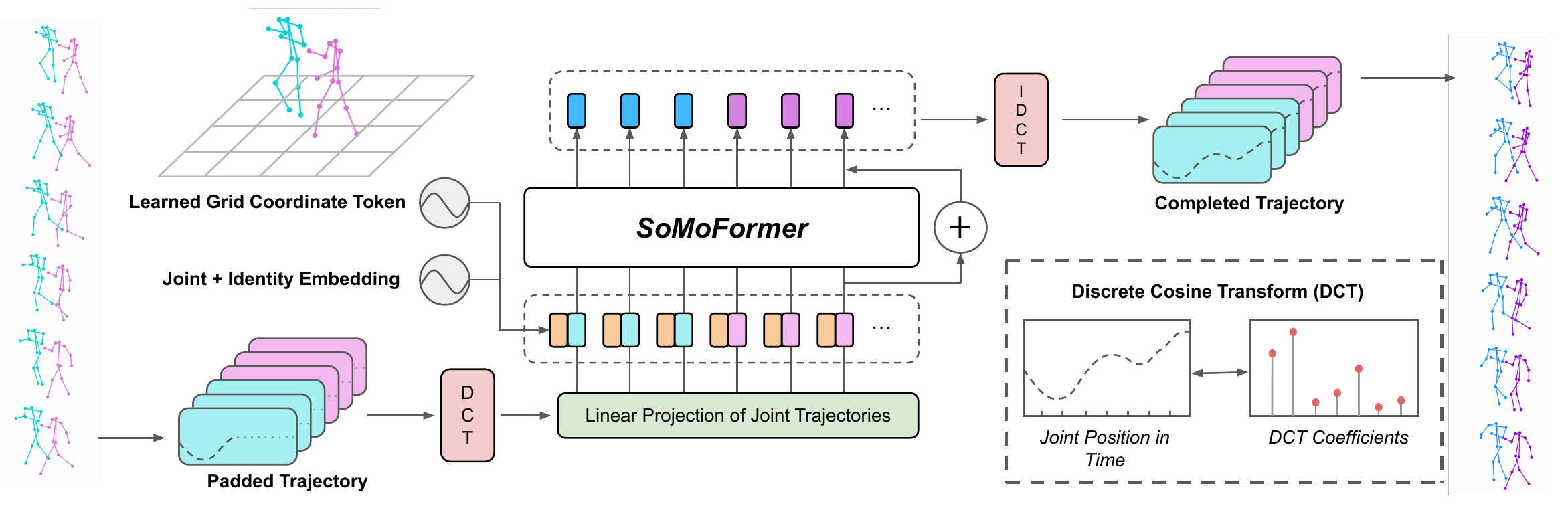}
    \caption{SoMoFormer uses a Transformer Encoder, which accepts each input token as the DCT-encoded (discrete cosine transform) padded trajectory of one joint. SoMoFormer learns a embedding for joint type and person identity which are added to the token, and then a grid position embedding encoding global position for each person is concatenated. SoMoFormer is able to predict pose trajectories for multiple people simultaneously while leveraging attention over joints to model human body dynamics.}
    \label{fig:arch}
\end{figure*}

To our knowledge, only \cite{wang2021multi} propose a multi-person Transformer for pose forecasting. Their approach uses both a local and global encoder to model temporal and social dependencies respectively, which limits them to predict poses just one person in the scene at a time, requiring multiple passes to produce a multi-person scene. This limitation affects all prior work including those that use social pooling methods. Furthermore, \cite{wang2021multi} address the problem of a variable number of people by using a global encoder with one input query for each frame for each person, which then introduces the challenge of multiple input queries corresponding to the same timestep. Indeed, all prior attention-based methods structure pose as a time series input, making extensions to multi-person cases difficult because they require the model to predict over an input containing both \textit{time}-series and \textit{person}-series sequences. 

In this work, we introduce SoMoFormer, an attention-based pose forecasting model which unifies multi-person pose forecasting (See Figs.~\ref{fig:visual} and \ref{fig:arch}). Different from the previous methods, we reformulate the problem of pose forecasting from predicting a time-sequence series of poses to predicting a joint-sequence series of trajectories where each input query denotes the trajectory of a skeleton joint coordinate over the entire input timeframe. Our re-framing of the pose forecasting problem allows us to design SoMoFormer, which accepts a set of skeleton joint trajectories as input rather than a set of pose timesteps, enhancing interactions among joints and allowing our model to predict an entire future sequence without recurrence. Crucially, this architecture naturally extends to multi-person scenes by simply adding more joint trajectories to the input query set for all people in the scene. SoMoFormer enables this new type of motion prediction by using a learned embedding to denote which person each joint belongs to as well as learned embeddings for joint type and location within the scene, to enable the prediction of a future multi-person pose sequence in one pass. Experimental results on multiple datasets show that SoMoFormer is superior at learning the relationships between joints and between people, and consequently outperforms all previous state-of-the-art methods in human pose forecasting. In summary, our contributions include:
\begin{itemize}
    	\setlength{\itemsep}{1pt}
	\setlength{\parskip}{0pt}
	\setlength{\parsep}{0pt}
    \item We introduce a novel transformer-based method, named SoMoFormer, which uses joint-aware attention for multi-person human pose forecasting.
    \item We reformulate the problem of pose forecasting by accepting and predicting a query for each joint rather than a query for each timestep, allowing SoMoFormer to perform long-term pose forecasting for multiple people without recurrence.
    \item SoMoFormer outperforms state-of-the-art methods for long-term motion prediction on the SoMoF benchmark \footnote{The SoMoF benchmark: \url{https://somof.stanford.edu/}}, as well as on the CMU-Mocap \cite{AMASS:ICCV:2019} and MuPoTS-3D \cite{singleshotmultiperson2018mupots} datasets.
    \item Analysis of our model attention weights reveals that SoMoFormer learns rich representations of human motion and dynamics.
\end{itemize}

\section{Related Work}

{\bf Human Trajectory Forecasting:}
Many recent works model human motion as a trajectory, representing each person as a 2-dimensional point on a plane \cite{Kosaraju2019SocialBiGATMT,Giuliari2021TransformerNF,Sadeghian2019SoPhieAA,Li2020EndtoendCP,Yu2020SpatioTemporalGT,Liu2021MultimodalMP}. \cite{Sadeghian2019SoPhieAA} use RNNs combined with an attentive GAN to provide socially plausible output predictions. \cite{Kosaraju2019SocialBiGATMT} use a graph attention network combined with a GAN to encode the social interactions between humans in the scene. Catalyzed by the success of the Transformer model \cite{Vaswani2017AttentionIA}, many recent works have used attention-based methods, such as transformers, to better model sequential data and complex human motion relationships \cite{Mao2019LearningTD,Mao2020HistoryRI,Mao2021MultilevelMA,Adeli2021TRiPODHT,Giuliari2021TransformerNF,Liu2021MultimodalMP,Adeli2021TRiPODHT,MartinezGonzalez2021PoseT}. Within trajectory forecasting, \cite{Giuliari2021TransformerNF} used a transformer to model the trajectory of each person through time. \cite{Liu2021MultimodalMP} improved upon this by proposing a stacked transformer that is able to use the motion from multiple objects in the scene. Nevertheless, trajectory forecasting methods fail to capture the complex pose dynamics involved in human motion.

{\bf Single-Person Pose Forecasting:}
Single-person human pose forecasting methods predict the future joint coordinates without global translation. Early single-person pose forecasting models employed Hidden Markov Models \cite{markov}, linear dynamics models \cite{lineardynamics}, and hand-crafted features such as hard constraints on limb length and joint angles \cite{adeli2012model,hierarchicalskeletaljoints}. Later methods moved toward end-to-end neural networks with few or no hand-crafted features. Many of these recent methods use an RNN backbone, with some using graph attention networks or GANs to extend the prediction to multiple entities or provide a plausible output. Chiu et al. \cite{Chiu2019ActionAgnosticHP} use a hierarchical RNN to predict human motion. \cite{Mao2019LearningTD} models human motion as a graph of encoded motions for each joint coordinate with a GNN architecture to pass information between nodes. Wang et al. \cite{Wang2021SimpleBF} extend this work by compressing the node representation to all dimensions of a joint. Mao et al. \cite{Mao2020HistoryRI} introduce an attention-based model to capture the similarities between current and historical motion sub-sequences, allowing the model to aggregate past motions for long-term prediction. Cai et al. \cite{jointpropagation} use a human skeleton model to propagate joint predictions based on kinematic connections. Matínez-González et al. \cite{MartinezGonzalez2021PoseT} used an encoder-decoder architecture to simultaneously decode all future timesteps without recurrence. All of the above methods remove global translation, modeling just local human skeletal motion. However, in a multi-person setting, such as the environment around an autonomous vehicle, global interactions between people must also be taken into account when making a prediction.

{\bf Multi-Person Pose Forecasting:}
With the need to model scenes with multiple people, some recent works focus on multi-person forecasting to predict the future for an entire scene. As with both pose and trajectory forecasting, these works similarly adopted attention-based methods. Adeli et al. \cite{Adeli2021TRiPODHT} use graph attentional networks to model interaction between humans and objects, but use an RNN to predict future motion. Matínez-González et al. \cite{MartinezGonzalez2021PoseT} use a transformer to predict an entire future sequence without recurrence, but use time-series queries making multi-person prediction infeasible. Wang et al. \cite{wang2021multi} use a transformer-based architecture to consider the global interactions between multiple people, but can only make inferences for one person at a time. Additionally, each input token for Wang et al. represents the position of all joints at a single timestep, making an extension to multi-person modeling unnatural. We propose a method to perform unified multi-person pose forecasting without recurrence by performing attention over a set of joints rather than a set of timesteps. Our method is able to use the attention mechanism of a transformer to learn the relationships between joints and between people simultaneously to naturally model human body dynamics and multi-person interactions.

\section{Methods}

Given a scene with $N$ humans in motion, each composed of $J$ joints, we wish to predict their motion for some number of future timesteps. Formally, for each person we are given the pose history sequence $X^n_{1:t} = [x^n_1, x^n_2, \ldots, x^n_t]$ where $n = 1, \ldots, N$ denotes the person, $t$ represents the input pose sequence length, and each $x_k^n = (j_1,j_2,\ldots,j_{J}) \in \mathbb{R}^{J \times 3}$ represents each pose of person $n$ at time $k$ as the three-dimensional Cartesian coordinates of each joint, in global coordinates. We aim to build a model to predict motion for each person for $T$ timesteps into the future, predicting the pose sequence $S^n_{t+1:T}$. 

\subsection{Proposed Architecture}

Figure \ref{fig:arch} is an overview of our proposed pose forecasting framework. Our model is a transformer encoder that accepts a token for each coordinate ($x$, $y$, and $z$) of every joint in the scene, encoded as the padded coordinate sequence over the input timeframe and transformed via DCT. Our model performs trajectory completion for $(X^n_{1:t}||\hat{X}^n_{t+1:T})$, where $||$ denotes concatenation, giving us the concatenation of the known history sequence $X^n_{1:t}$ and padded sequence $\hat{X}^n_{t+1:T}$ generated using the previous history. By using a padded sequence, our input and output queries are the same length, allowing the model to predict the difference between the DCT representation of the input and output instead of generating a full DCT sequence. We use a constant-padded sequence rather than a random or fixed padding value because seeding the output with a reasonable initialization makes the prediction task simpler. Taking advantage of residual connections within the transformer, our model refines the predicted motion at each layer. Differing from previous multi-person pose forecasting methods \cite{wang2021multi} our model accepts one input query for each joint, rather than for each timestep. Thus our model accepts $N\times J \times 3$ input queries in total for a scene with $N$ people, where each input query is a vector of size $(t+T)$ containing one cartesian dimension of a joint trajectory over all $(t+T)$ timesteps. 

To improve generalization to large scenes or distant people, our model accepts joint queries with global translation removed by subtracting the neck or pelvis joint coordinate of the last-known ($t$-th) frame for every person. This translation is added back afterward to recover the global coordinate position. Since the input queries for each person do not have global translation information, the global position of each person is instead encoded via a learned embedding, and concatenated with the joint trajectory representation. When a scene has multiple people, our model accepts queries for all people in a scene. We encode both joint type (left knee, right shoulder, etc.) and personal identity with learnable embeddings which are added to the joint queries.

\subsection{Discrete Cosine Transform}

\begin{figure*}[t]
    \centering
    \includegraphics[width=1.0\linewidth]{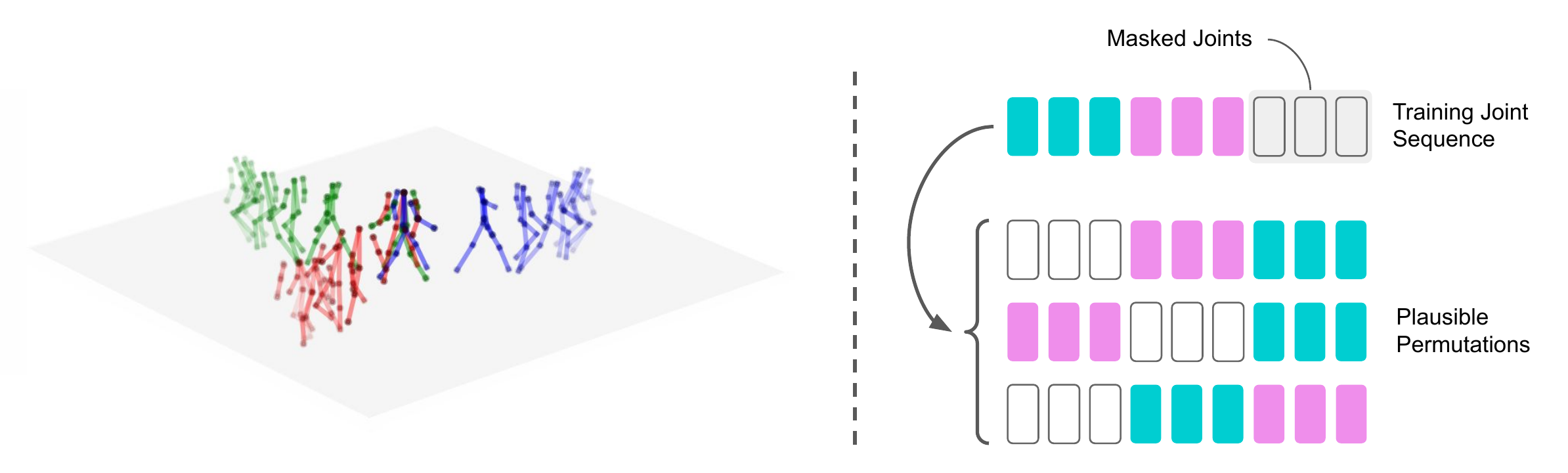}
    \caption{Left: We randomly rotate poses around the y-axis at training time. The red, blue, and green pose trajectories are all rotations of the same pose sequence. Right: We randomly permute the order of people so that the learned identity embedding sees a more diverse set of training examples. Each color represents joints from a different person or a masked query. We keep joints from the same person together so that the identity embedding is consistent.}
    \label{fig:rotation}
\end{figure*}

Predicting Cartesian coordinates directly does not take advantage of human motion being continuous, smooth, and somewhat periodic. Instead, recent works \cite{Mao2020HistoryRI,wang2021multi} use a Discrete Cosine Transform (DCT) to encode human motion into the frequency domain as a collection of coefficients. In addition to more reasonably encoding human motion, using DCTs allows us to predict the entire trajectory for all future frames simultaneously. 

For a motion sequence of a single coordinate $(x_1, \ldots , x_T)$, the $l$-th DCT coefficient is computed by
\begin{align}
    C_l = \sqrt{\frac{2}{T}}\sum_{t=1}^T \frac{x_t}{\sqrt{1 + \delta_{l1}}}cos\frac{\pi}{2T}(2t-1)(l-1)
\end{align}
for
\begin{align}
    \delta_{ij} = \begin{cases}
                                   1 & \text{if $i=j$} \\
                                   0 & \text{if $i\neq j$}
  \end{cases}
\end{align}
where $l = 1, 2, \ldots, T$, yielding the same number of coefficients as the sequence length. This process gives the coefficient sequence $(C_1, \ldots , C_T)$ which is used as an input token to the model for the corresponding joints coordinate to predict the final, completed trajectory $(\tilde{C}_1, \ldots , \tilde{C}_T)$. Given this coefficient sequence, we can recover the coordinates using the inverse DCT, where we get a specific coordinate via
\begin{align}
    \tilde{x}_t = \sqrt{\frac{2}{T}}\sum_{l=1}^T \frac{\tilde{C}_l}{\sqrt{1 + \delta_{l1}}}cos\frac{\pi}{2T}(2t-1)(l-1).
\end{align}

In practice, this DCT encoding is easy to use as it can be implemented with a simple invertible linear transformation and significantly improves results.

\subsection{Person and Joint Embeddings}

Since the input into a Transformer is position invariant, Transformers use embeddings to denote the input query order, generally in the form of a vector that is added or concatenated with the input query. In natural language processing and other time series methods, a common positional embedding include alternating pairs of sines and cosines to denote the input order \cite{devlin-etal-2019-bert}. Similarly, for each input joint query, our model should encode both its joint type (neck, hip, shoulder, etc.) as well as the identity of the person it corresponds to. We accomplish this by introducing the use of two distinct embeddings denoting joint type and person identity. For each input joint query, we add both an embedding corresponding to the joint type, and an embedding corresponding to the person's identity. Both embeddings are randomly-initialized and learned at training time.

\subsection{Grid Positioning}

Since we subtract global translation in the model input queries for easier training, we use a grid positioning method to provide global translation information to the model. Inspired by the Vision Transformer \cite{dosovitskiy2020vit} which divides an image into a patch grid and encodes patch placement with positional embedding, we divide the global scene into a grid of cells and assign each cell a randomly-initialized, learnable positional embedding, which is concatenated to the input query. Each person in a scene is assigned to a cell based on the position of the neck joint at the last known frame. The model learns to associate the grid embedding to the distance between different people in a scene.

\subsection{Random First-Frame Selection}

Our training datasets often consist of extended scenes with many frames. We sample a subsequence from the entire longer sequence by randomly selecting a start frame and taking the pose trajectory over the next $t+T$ frames, which increases the diversity of training examples as compared to simply dividing each scene into fixed subsequences. Like in \cite{Wang2021SimpleBF}, we further augment our data by using reversed pose sequences as additional training data.

\subsection{Random Rotation}

At training time, we randomly rotate scenes about the vertical axis on a uniform distribution within $[0, 2\pi]$, creating scenes that are equally plausible. This data augmentation method helps prevent the model from overfitting to a specific trajectory. The left diagram in Figure \ref{fig:rotation} visualizes a pose sequence under several rotations. Note that each of the rotated pose sequences is valid and useful. Our use of random rotation is crucial to the effectiveness of a learned grid embedding because it helps expose every grid cell to a variety of scenes.

\subsection{Random Person Permutation}

Since the order in which people are input into the model should not matter, we randomly permute this order at training time, as described in Figure \ref{fig:rotation}, to make our learned identity embedding more robust. Our training data consists of scenes with different numbers of people, requiring us to use a padding mask with our Transformer model. In these cases, we also permute non-padded people across the elements of the padding mask as depicted in the figure. This technique helps prevent overfitting in which the model could associate specific poses with specific identity embeddings. Additionally, by permuting padded examples, we allow all identity embedding to see an equal share of non-padded training examples.

\section{Experiments}
This section details the experiments we performed to evaluate our approach.

\begin{table*}\small
    \centering
    \caption{Experimental results in VIM on the SoMoF 3DPW test set.}
    \renewcommand{\arraystretch}{1.1}
    \begin{tabular}{lllllll}
     \hline
                                        & \multicolumn{5}{c}{3DPW Prediction in Time}                               \\  [0.2ex] \cline{2-7} 
    Method                              & 100ms & 240ms   & 500ms           & 640ms           & 900ms     & \textbf{Overall}      \\  [0.2ex] \hline
    Mo-Att \cite{Mao2020HistoryRI} + ST-GAT \cite{huangstgat} & 
        62.4 & 94.6 & 153.2 & 188.0 & 249.9 & 149.6 \\
    SC-MPF \cite{Adeli2020SociallyAC}
        & 45.4   & 73.7 & 129.2 & 159.5 & 208.3 & 123.2 \\
    Zero Velocity
        & 29.4 &	53.6 &	94.5 &	112.7 &	143.1 & 86.7        \\
    TRiPOD \cite{Adeli2021TRiPODHT}
        & 31.0  & 50.8  & 84.7   & 104.1  & 150.4 & 84.2         \\
    DViTA \cite{Parsaeifarddecoupled}  & 19.5         & 36.9          & 68.3          & 85.5          & 118.2 & 65.7         \\
    FutureMotion \cite{wang2021multi}                        & 9.5          & 22.9          & 50.9          & 66.2          & 97.4 & 49.4         \\
    \textbf{SoMoFormer} & \textbf{9.1} & \textbf{21.3} & \textbf{47.5} & \textbf{61.6} & \textbf{91.9} & \textbf{46.3} \\ \hline
    \end{tabular}
    \label{table:somof_benchmark}
\end{table*}

\begin{table*}\small
    \centering
    \caption{Experimental results in MPJPE on CMU-Mocap and MuPoTS-3D.}
    \renewcommand{\arraystretch}{1.1}
    \begin{tabular}{llll|lll}
     \hline
    & \multicolumn{3}{c}{CMU-Mocap Test Set} & \multicolumn{3}{c}{MuPoTS-3D Test Set}                      \\  [0.2ex] \cline{2-4} \cline{5-7} 
    Method                              
            &1 sec \ \ 
            &2 sec \ \  
            &3 sec \ \  
            &\ \ 1 sec \ \  
            &2 sec \ \  
            &3 sec \ \    \\  [0.2ex] \hline
HRI \cite{Mao2020HistoryRI} & 0.503 & 0.932 & 1.422 & \ \ 0.261 & 0.469 & 0.714 \\

LTD \cite{Mao2019LearningTD}   & 0.480 & 0.869 & 1.181 &  \ \ 0.191 & 0.337 & 0.466 \\

    Multi-Range \cite{wang2021multi} \ \  &
0.456 & 0.840 & 1.11 & \ \ 0.206 & 0.393 & 0.574 \\

    \textbf{SoMoFormer} &
\textbf{0.42} & \textbf{0.80} & \textbf{1.06} & \ \ \textbf{0.173} & \textbf{0.305} & \textbf{0.423}
          \\ \hline
    \end{tabular}
    \label{table:multirange}
\end{table*}

\subsection{Metrics}

\noindent
\textbf{MPJPE. } Consisted with prior work \cite{MartinezGonzalez2021PoseT,Mao2020HistoryRI,wang2021multi}, we report our Mean Per Joint Position Error (MPJPE), which measures the average Euclidean distances between ground-truth and predicted joint positions. We use global coordinates to better evaluate when the predicted trajectory deviates from the ground truth.

\noindent
\textbf{Visibility-Ignored Metric (VIM). } First introduced in \cite{Adeli2021TRiPODHT}, VIM calculates the mean $3J$-dimensional distance between the ground-truth and predicted joint positions after flattening together the joint and coordinate dimensions. This is the metric used for the SoMoF Benchmark.

\subsection{Data}
We use the following datasets to train our model. These datasets provide human body meshes, which are regressed to produce joint coordinates.

\noindent
\textbf{3DPW \cite{vonMarcard20183dpw}. } We used the 3D Poses in the Wild Dataset (3DPW) in our experiments for human motion prediction. This dataset provides human motion sequences in real-world settings, with over 60 video sequences. Since we used the SoMoF benchmark to evaluate our model, we use the SoMoF benchmark splits for 3DPW in which the 3DPW train and test set are flipped. This means that we technically train our model using the 3DPW test set, and evaluate on the 3DPW train set.

\noindent
\textbf{AMASS \cite{AMASS:ICCV:2019}.} The Archive of Motion Capture As Surface Shapes (AMASS) provides a massive dataset of human motion capture sequences with over 40 hours of motion and 11,000 motions provided as SMPL mesh models. During training, we use the \textit{CMU}, \textit{BMLMovi}, and \textit{BMLRub} subsets of this dataset, which provide large-scale and varied sets of motions. Since many of these sequences are single-person, we synthesize additional training data by mixing together sampled sequences to create multi-person training data. 

\noindent
\textbf{CMU-Mocap}. The Carnegie Mellon University Motion Capture Database (CMU-Mocap) provides motion capture recording from 140 subjects performing various activities. We use training and testing sets derived by Wang et al. \cite{wang2021multi} to train and evaluate our model.

\noindent
\textbf{MuPoTS-3D \cite{singleshotmultiperson2018mupots}}. The Multi-person Pose estimation Test Set in 3D (MuPoTS-3D) provides 8,000 annotated frames of poses from 20 real-world scenes. We use this test dataset to evaluate our model performance. 

\subsection{Training}

\begin{table*}[t]\small
    \centering
    \caption{Ablation study on the SoMoF 3DPW validation set, with results shown in VIM (top) and MPJPE (bottom). Note that this 3DPW validation set is different from the 3DPW test set in Table 1.}\vspace{.2em}
    \renewcommand{\arraystretch}{1.1}
    \begin{tabular}{lcccccc}
     \hline
    Method & 100ms & 240ms & 500ms & 640ms  & 900ms & \textbf{Overall}  \\ [0.2ex] \hline

    Baseline \ \  &
        9.0 &  23.3 &  54.5 & 68.9 & 95.8 & 50.3 \\
    + Learned Embedding \ \  &
        8.6 &  22.0 &  50.3 & 63.6 & 92.3 & 47.4 \\
    + Data Augmentation \ \  &
        8.0 &  20.3 &  46.2 & 58.4 & 84.3 & 43.5 \\
    + Grid Positioning \ \  &
        \textbf{7.6} &  \textbf{19.7} &  \textbf{44.6} & \textbf{55.9} & \textbf{80.2} & \textbf{41.6} \\
    
    \hline \hline
    
    Baseline \ \  &
        2.0 &  5.2 &  12.7 & 16.4 & 23.6 & 12.0 \\
    + Learned Embedding \ \  &
        1.9 &  4.9 &  11.9 & 15.6 & 23.4 & 11.5 \\
    + Data Augmentation \ \  &
        1.8 &  4.5 &  11.0 & 14.4 & 21.4 & 10.6 \\
    + Grid Positioning \ \  &
        \textbf{1.7} &  \textbf{4.4} &  \textbf{10.6} & \textbf{13.7} & \textbf{20.2} & \textbf{10.1}
          \\ \hline
    \end{tabular}
    \label{table:somof_ablation}
\end{table*}
Rather than simply taking the loss between the outputs of the transformer and the ground truth, we further encourage our model to avoid overfitting and learn meaningful motion representations at each layer by taking an auxiliary loss on the output of every layer. Intuitively, this allows the model to iteratively refine the predicted motion at each layer. Given a dataset with $M$ training examples and an $L$-layer, $S^n_{1:t+T} \in \mathbb{R}^{Jx3}$ denotes the ground-truth 3D coordinates and$S^n_{\ell, 1:t+T}$ denotes the predicted 3D coordinates at the $\ell$-th layer. We use $L_2$ loss to minimize the error between the ground truth and predicted coordinates:
\begin{align}
    \mathcal{L} = \sum_{\ell=1}^L\sum_{i=1}^M\sum_{n=1}^{N} \lambda_L \|S^n_{t+1:t+T} - \hat{S}^n_{L, t+1:t+T}\|^2
\end{align}
where $\lambda_\ell$ denotes the weight of the loss from the $\ell$-th layer.

\subsection{Implementation Details}
Our transformer encoder has 6 layers, 8 attention heads, and a hidden dimension of 1024. We train for 256 epochs with a batch size of 512. We use the Adam optimizer with an initial learning rate of 0.001, which is decayed by 0.1 at the end of training. We use $\lambda_6 = 1, \lambda_{1...5} = 0.2$. We encode global position using a grid embedding by dividing the scene into 5x5 grid, corresponding to 25 unique grid embeddings. SoMoFormer is developed using PyTorch.

\section{Experimental Results}

We first show that our method outperforms previous approaches in human pose forecasting on the SoMoF 3DPW, CMU-Mocap, and MuPoTS-3D test sets. Then, we perform an ablation study to provide insight into our model and training methods. Using visualized prediction, we qualitatively validate the performance of our model. Finally, we visualize the attention graphs from our model to demonstrate that it learns meaningful human body motion representations.

\subsection{Results on SoMoF Benchmark}
The SoMoF benchmark \cite{Adeli2020SociallyAC,Adeli2021TRiPODHT} provides a benchmark for multi-person human pose trajectory. Each sequence has 16 frames (1070 ms) of input to predict the next 14 frames (930 ms), where each frames consists of joints positions for multiple people. Results are reported as the mean VIM at multiple future timesteps. Just as \cite{Wang2021SimpleBF}, we train on the 3DPW \cite{vonMarcard20183dpw} and AMASS \cite{AMASS:ICCV:2019} datasets, which provide both multi-person and single-person data, and finally we finetune on 3DPW. Since SoMoF only uses 13 joints for evaluation, we use just these joints during training as well. We report in Table \ref{table:somof_benchmark} a comparison of methods on the SoMoF 3DPW test set \footnote{Our submission (currently anonymous) to the SoMoF benchmark is dated March 7, 2022.}. Our model consistently outperforms all previous methods.

\subsection{Results on CMU-Mocap and MuPoTS-3D}

We additionally compare our method to Wang et al. \cite{wang2021multi} who proposed a recent method for multi-person pose forecasting achieving state-of-the-art results on several datasets, as well as other recent methods including HRI \cite{Mao2020HistoryRI} and LTD \cite{Mao2019LearningTD}. Using their protocols, models are trained on a synthesized dataset mixing sampled motions from CMU-Mocap to create 3-person scenes, and evaluated on both CMU-Mocap and MuPoTS-3D. We give 15 frames (1000 ms) of history as input to predict the next 45 frames (3000 ms), and report the MPJPE at 1, 2, and 3 seconds in the future. For fair comparison, we train and evaluate each method using the code and data provided by \cite{wang2021multi}. We report in Table \ref{table:multirange} a comparison of each method on the testing datasets. Our model consistently outperforms the others on both CMU-Mocap and MuPoTS-3D. We observe a significant difference in performance on MuPoTS-3D, which unlike CMU-Mocap is not synthesized from a combination of samples poses, meaning that the data represents real social interactions. While \cite{wang2021multi} predict the future pose forecast for just one person at a time, SoMoFormer is able to fully leverage attention over joints from all people in a scene to make predictions for all people simultaneously. When data contains interactions between multiple people, as in MuPoTS-3D, our method can better model these interactions to predict the future.

\begin{figure}[tb]
    \centering
    \includegraphics[width=\linewidth]{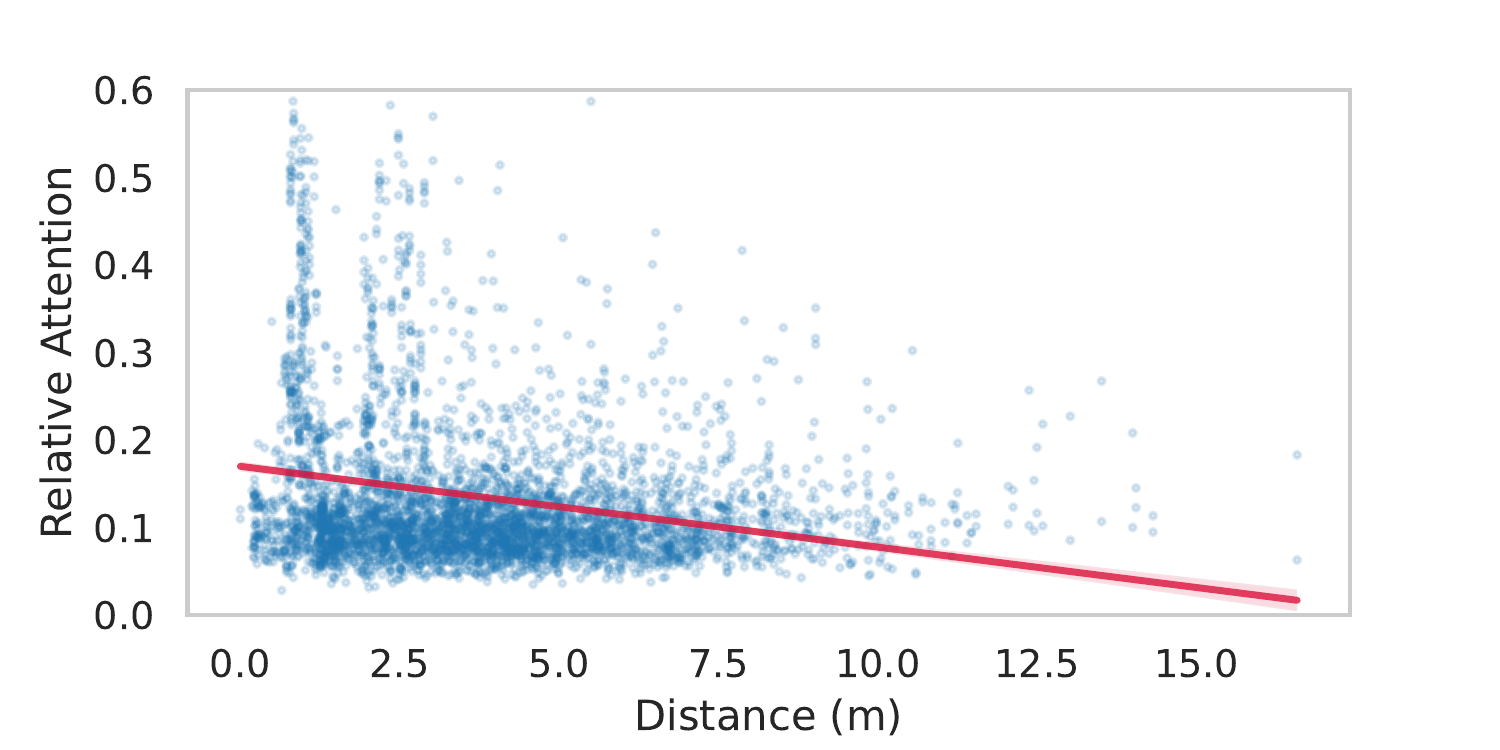}
    \caption{\footnotesize A plot of the distance between two people against the attention score in SoMoFormer, averaged across joints. Each sample (\textcolor{blue}{blue}) is drawn from 3-person scenes in CMU-Mocap, and we draw the best-fit line (\textcolor{red}{red}). The closer two people are, the higher the attention score between them. }
    \label{fig:attention_plot}
\end{figure}

\subsection{Ablation Study}

SoMoFormer introduces several components in order to achieve state-of-the-art performance: learned joint type and identity embeddings, data augmentation, and grid positioning. We perform an ablation study on each of these components to examine their contribution to SoMoFormer in Table \ref{table:somof_ablation}. Each model is trained on the SoMoF 3DPW training set and evaluated on the SoMoF 3DPW validation set in both VIM and MPJPE. We add each component one-by-one to show their effect on the final performance. The results indicate that each component benefits the performance of the model.

\noindent
\textbf{Learned Embeddings.} Learned joint embeddings allow our method to disambiguate joint types, just as positional embedding in time-series transformer models denote input order. The model can use these embeddings to learn relationships between types joints to better predict human motion dynamics.

\noindent
\textbf{Data Augmentation.} Our data augmentation methods include random first-frame selection, sequence flipping, and person permutation. These methods produce a wide variety of plausible scenes from a small dataset, allowing our model to generalize well to unseen examples without additional training data. Data augmentation gives a significant improvement in performance without architectural changes.

\noindent
\textbf{Grid Positioning.} We use an embedding to denote the position of all people within a scene, allowing our model to use global location information. In a multi-person setting, the location of people in a scene is useful for predicting social interactions. The attention mechanism of the transformer exploits the identity and global position of each person to make an effective multi-person prediction.

\subsection{Interactions between multiple people}
We can validate that SoMoFormer models the interactions between multiple people by observing the relationship between two people's distance and the average attention score imputed between their joint pairs by SoMoFormer.    Figure~\ref{fig:attention_plot} shows a clear relationship between two people's distance and the attention score, providing clear evidence that our model considers interactions between people in making a prediction.

\subsection{Visualized Examples}

We subjectively evaluate the performance of our model in Figure \ref{fig:visual} by visualizing predicted motion sequences on examples from the SoMoF 3DPW test set. Comparing the real and predicted sequences, we observe that our model's predictions closely match the ground-truth even in a multi-person motion scenario with two people walking side-by-side. These results validate the performance of our model.

\subsection{Attention Weights Visualization}

\begin{figure*}[t]
    \centering
    \includegraphics[width=0.25\linewidth]{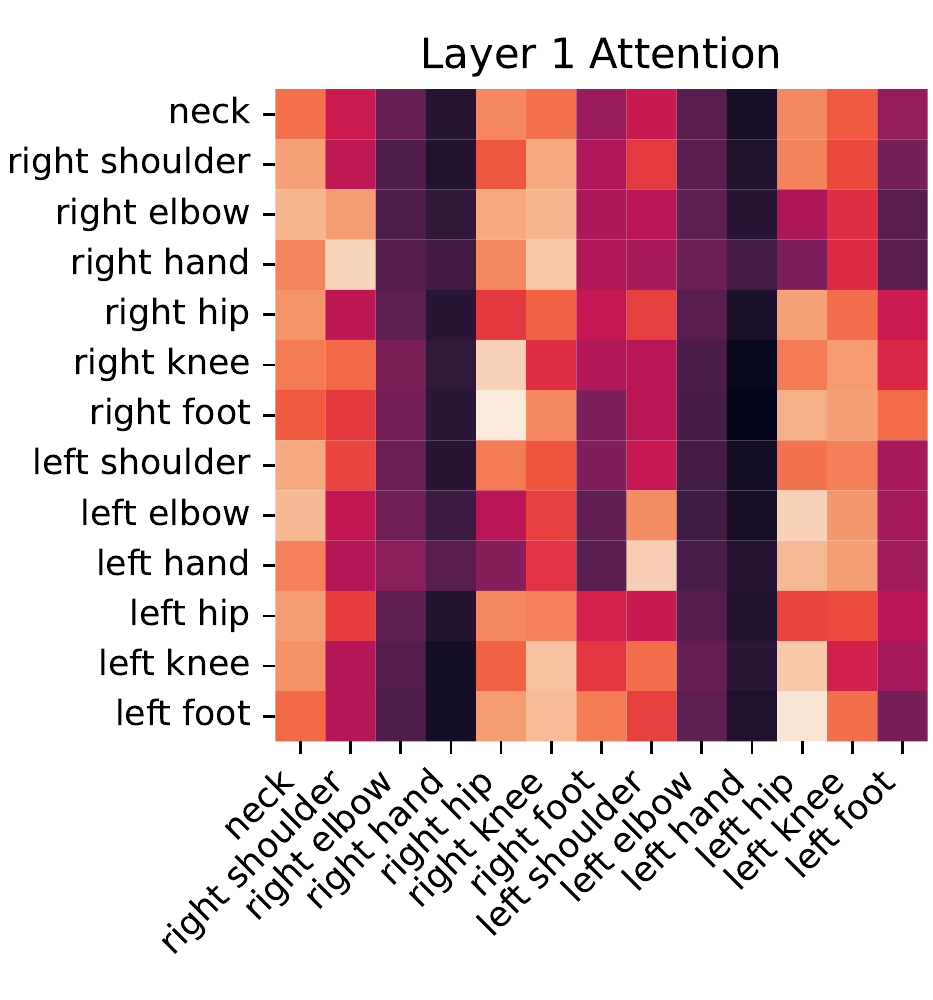}
    \includegraphics[width=0.25\linewidth]{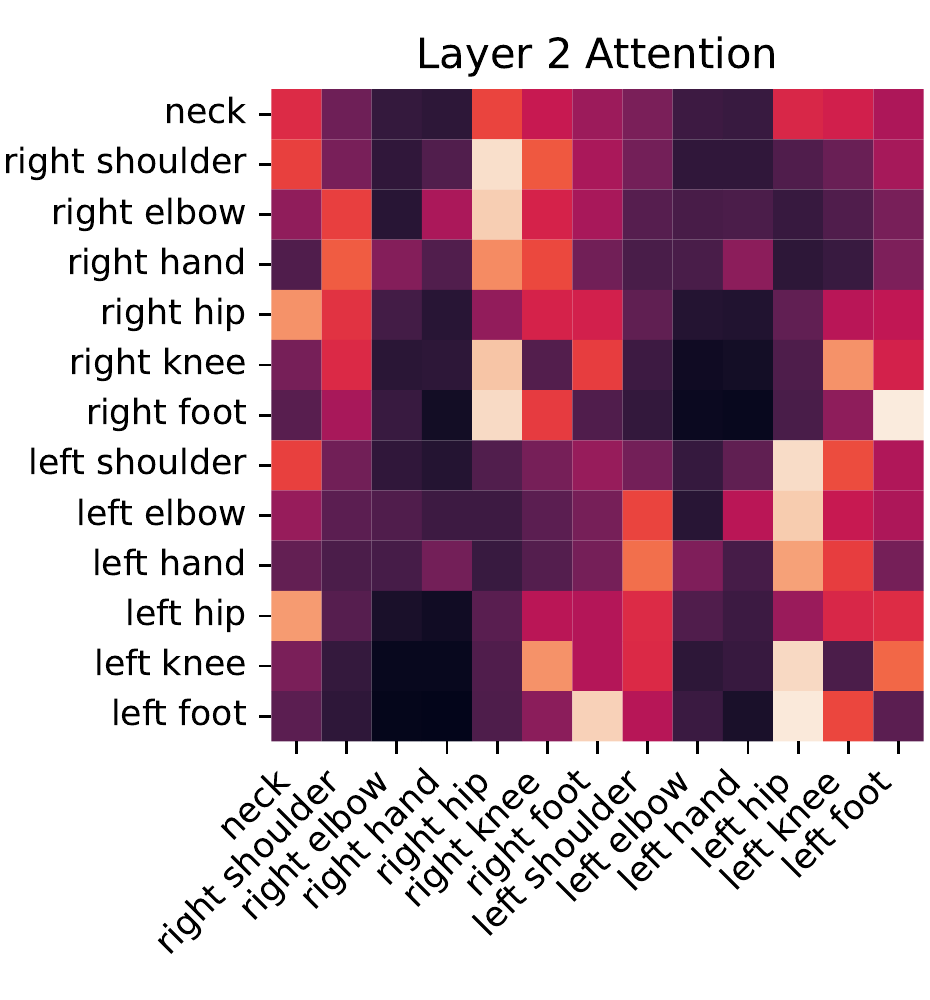}
    \includegraphics[width=0.25\linewidth]{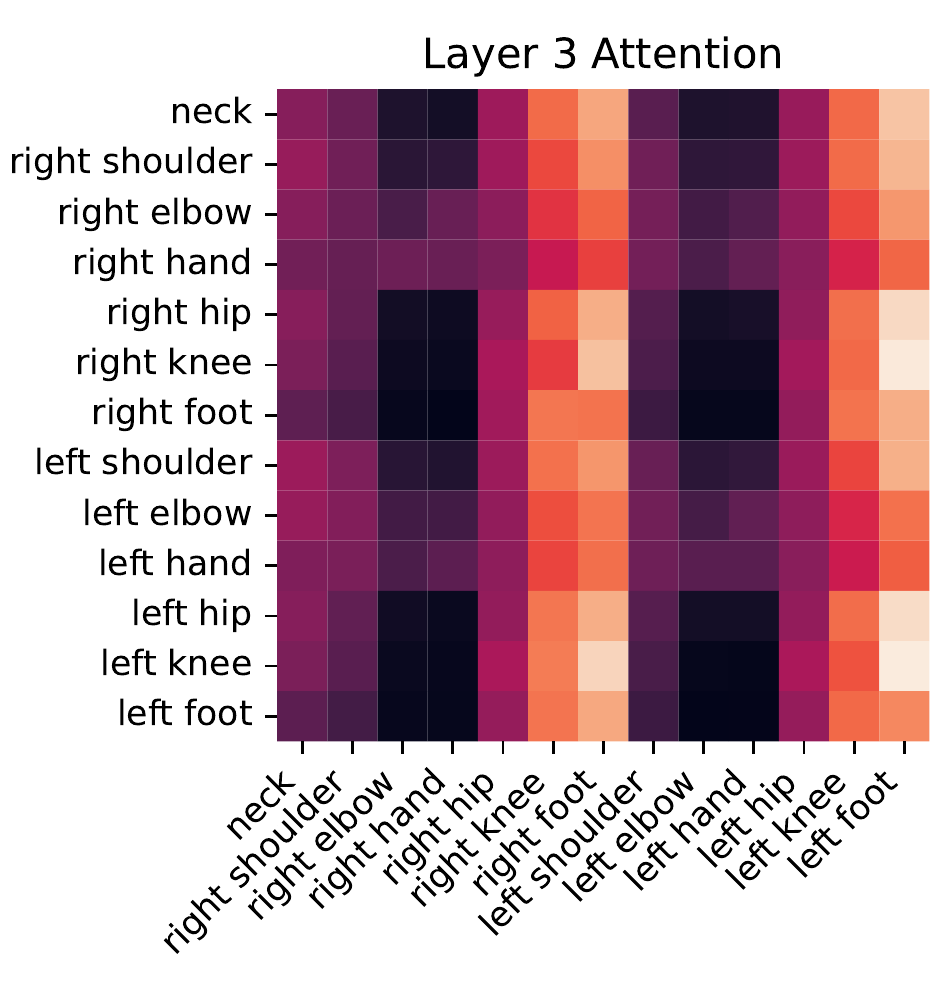} 
    \vspace{-1.1em}
    \\
    \includegraphics[width=0.25\linewidth]{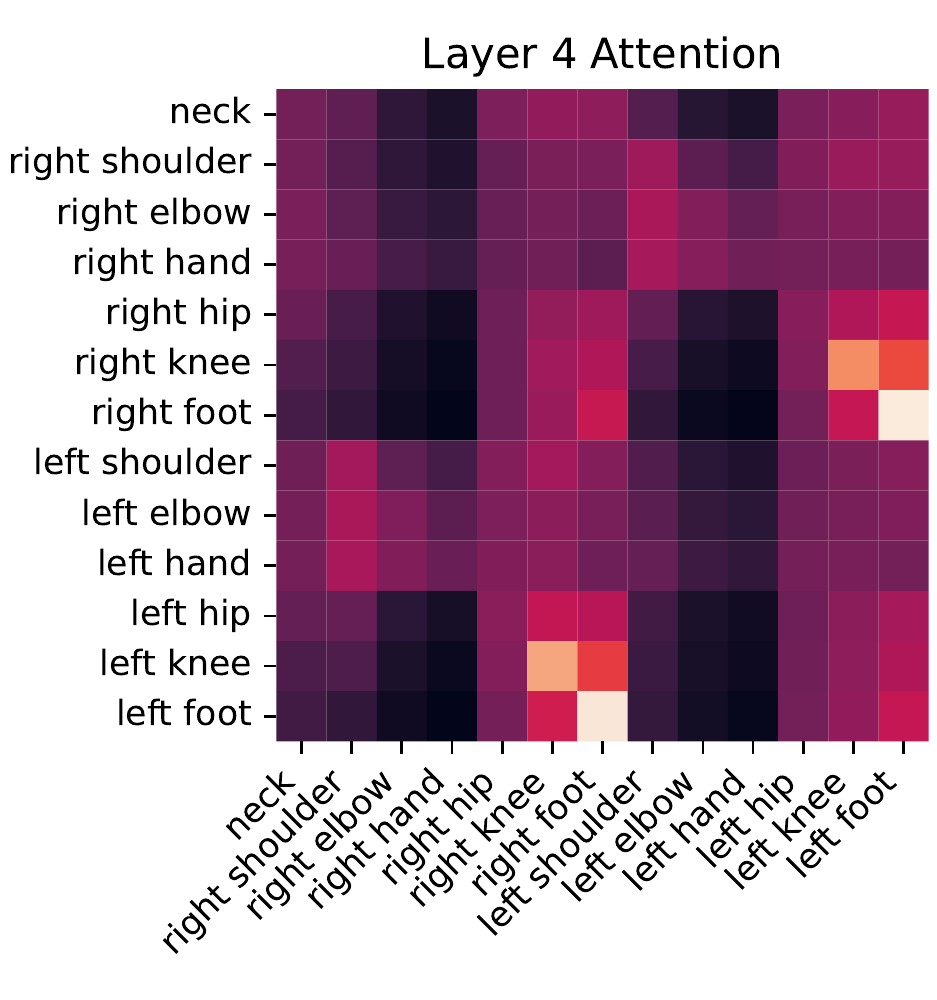}
    \includegraphics[width=0.25\linewidth]{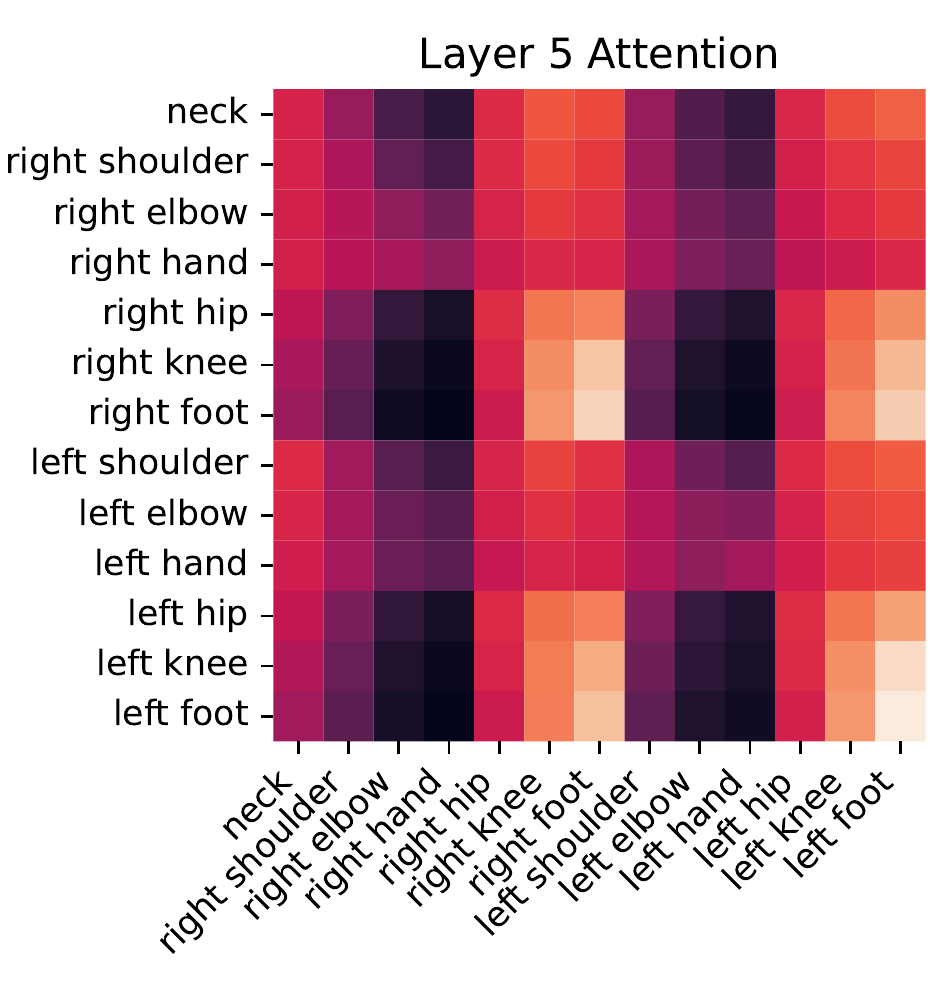}
    \includegraphics[width=0.25\linewidth]{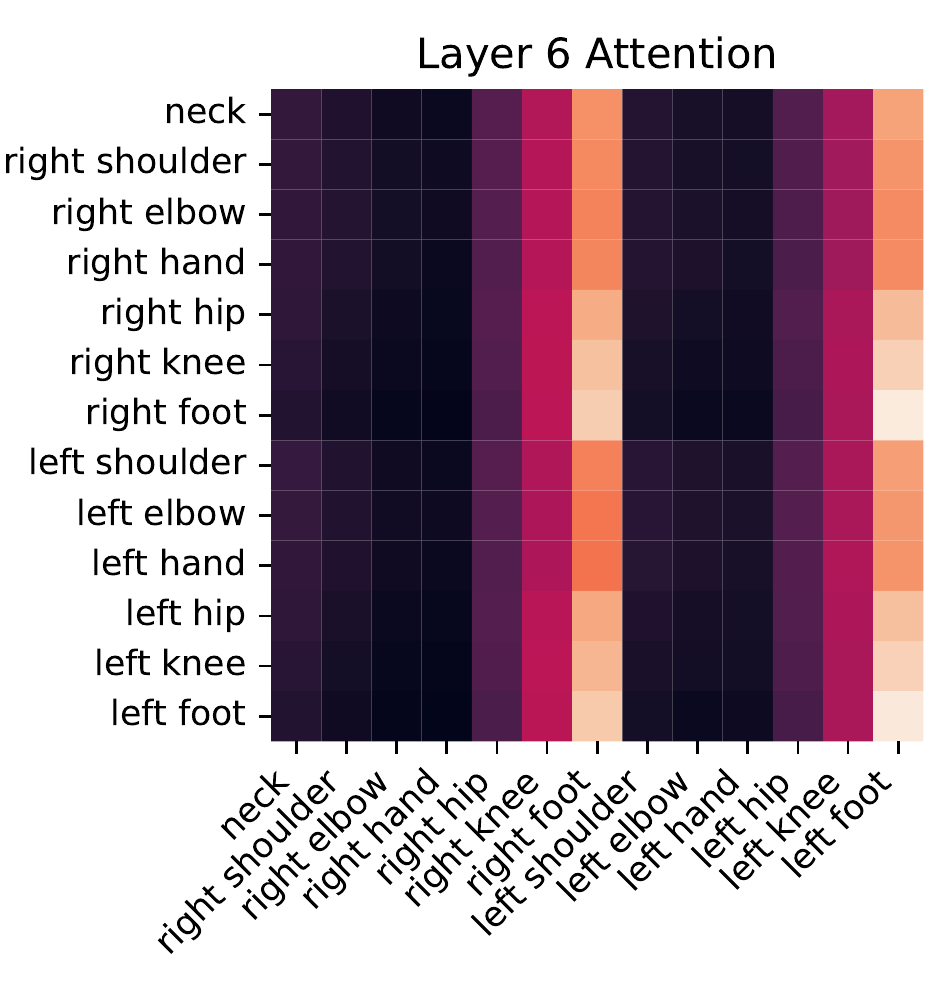}
    \caption{A visualization of attention scores between joint types at different layers of the model. Each row represents the attention scores from a single joint as inferred by SoMoFormer. A brighter color indicates a stronger attention score from the row joint to the column joint.}
    \label{fig:attention}
\end{figure*}

To further understand our method's ability to model interactions between joints, we visualize the attention weights of our model, as well as predicted attention of various joints in different motion scenarios. In Figure \ref{fig:attention} we visualize the average predicted attention weights across the entire SoMoF validation set. Figure \ref{fig:attention} shows that different layers attend to different input joints when making predictions, and these attention weights correspond to realistic human body dynamics. For example, in the Layer 2 attention, joints on the right side of the body attend strongly to the right hip joint, and vice-versa. Layers 4 and 5 show strong and symmetrical attention between leg joints. Interestingly, we note that the model learns to treat joints from the left and right sides of the body in similar ways, based on nearly identical attention scores, despite being given no explicit representation of a human skeleton. Overall, these results show that SoMoFormer leverages human body dynamics between joints to make accurate motion predictions, confirming the effectiveness of our joint-based approach.

\begin{figure}
    \centering
    \includegraphics[width=\linewidth]{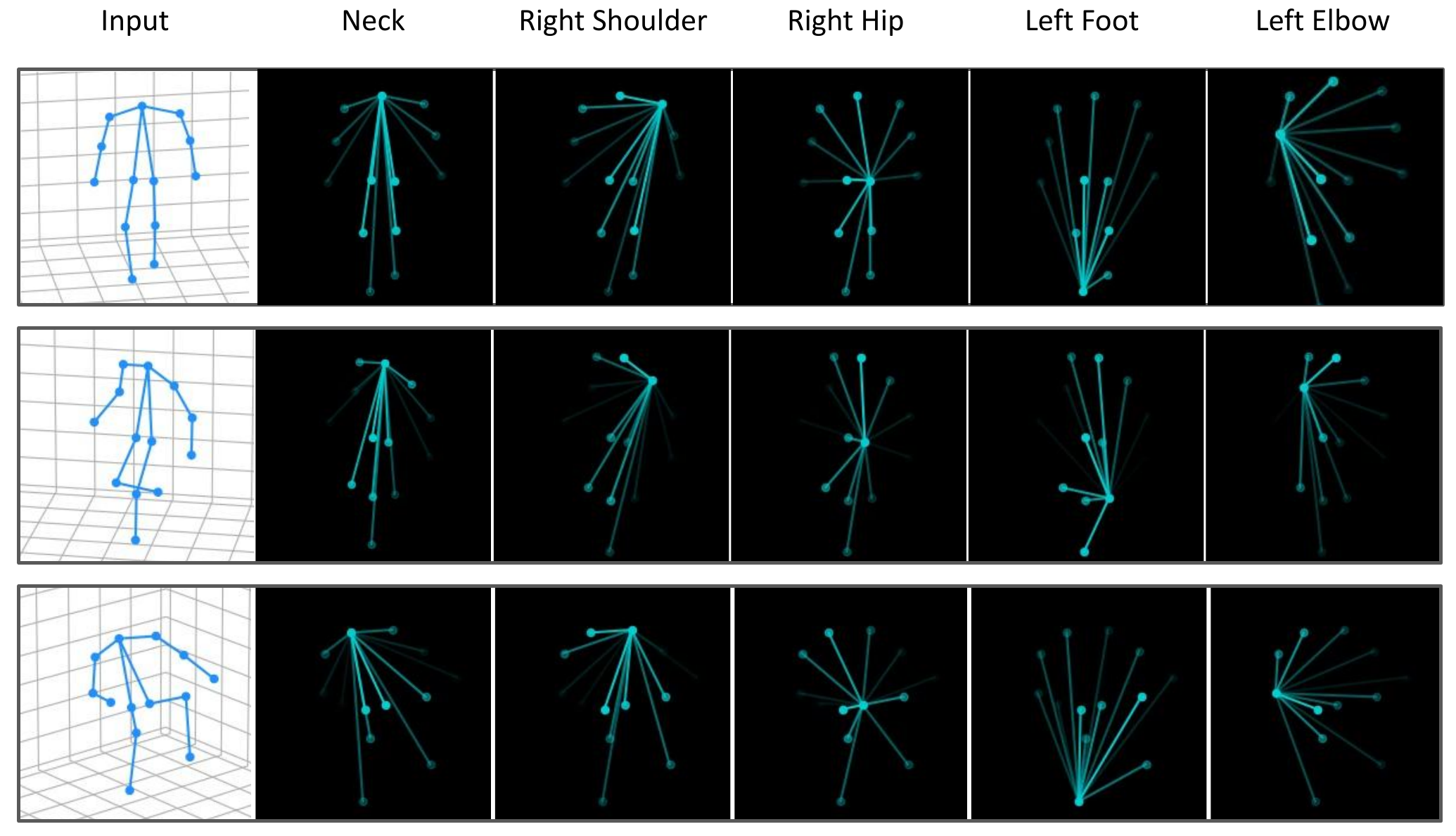}
    \caption{Qualitative results on the SoMoF dataset. We visualized self-attention scores from a joint to all other joints, where a brighter color indicates a stronger attention score. We observe that SoMoFormer learns a rich representation of human body motion between joints.}
    \label{fig:attention_joints}
\end{figure}

Inspired by \cite{metro}, in Figure \ref{fig:attention_joints} we analyze the attention scores for several motion examples by visualizing the scores between a joint and all other joints, where a brighter color indicates a stronger attention score. This visualization shows that (1) Joints tend to have higher attention to physically closer joints, and (2) In moving scenarios, joints tend to have higher attention scores to other joints that are most relevant to the motion. For example, we observe in the first standing scenario, the left foot joint attends strongly to other leg joints and more weakly to arm joints. However, in the second walking scenario, the left foot joints attend more strongly to the neck joint, which defines overall motion, and gives almost no attention to the arm joints which are not as relevant to the prediction. Figure \ref{fig:attention_joints} provides further evidence that our method learns a rich representation of human body motion. Further analysis would be needed to fully explore the relationships between joints in the model predictions.

\section{Conclusion}

In this paper, we proposed SoMoFormer, a novel framework for multi-person pose forecasting. Using attention over joint trajectories, our method predicts an entire future pose sequence for multiple people while modeling complex interactions among body joints. Experiment results show state-of-the-art performance on the SoMoF benchmark, as well as the CMU-Mocap and MuPoTS-3D datasets.

Our model can be used in a range of applications, from autonomous vehicle technology to realistic human motion for skeletal meshes. SoMoFormer can easily generalize to many people in a scene, and its efficiency in generating an entire human pose trajectory at once allows it to be used in low latency applications such as autonomous vehicles and gaming for realistic human motion. A detailed analysis reveals that SoMoFormer learns rich representations of human motion by modeling complex relationships between joints. We hope to further explore the semantic meaning and downstream applicability of these representations.

{\small
\bibliographystyle{splncs04}
\bibliography{eccvbib}

\begin{thebibliography}{10}
\providecommand{\url}[1]{\texttt{#1}}
\providecommand{\urlprefix}{URL }
\providecommand{\doi}[1]{https://doi.org/#1}

\bibitem{Adeli2020SociallyAC}
Adeli, V., Adeli, E., Reid, I.D., Niebles, J.C., Rezatofighi, H.: Socially and
  contextually aware human motion and pose forecasting. IEEE Robotics and
  Automation Letters  \textbf{5},  6033--6040 (2020)

\bibitem{Adeli2021TRiPODHT}
Adeli, V., Ehsanpour, M., Reid, I., Niebles, J.C., Savarese, S., Adeli, E.,
  Rezatofighi, H.: Tripod: Human trajectory and pose dynamics forecasting in
  the wild. In: Proceedings of the IEEE/CVF International Conference on
  Computer Vision. pp. 13390--13400 (2021)

\bibitem{adeli2012model}
Adeli-Mosabbeb, E., Fathy, M., Zargari, F.: Model-based human gait tracking, 3d
  reconstruction and recognition in uncalibrated monocular video. The Imaging
  Science Journal  \textbf{60}(1),  9--28 (2012)

\bibitem{markov}
Brand, M., Hertzmann, A.: Style machines. In: Proceedings of the 27th annual
  conference on Computer graphics and interactive techniques. pp. 183--192
  (2000)

\bibitem{jointpropagation}
Cai, Y., Huang, L., Wang, Y., Cham, T.J., Cai, J., Yuan, J., Liu, J., Yang, X.,
  Zhu, Y., Shen, X., et~al.: Learning progressive joint propagation for human
  motion prediction. In: European Conference on Computer Vision. pp. 226--242.
  Springer (2020)

\bibitem{Chiu2019ActionAgnosticHP}
kuang Chiu, H., Adeli, E., Wang, B., Huang, D.A., Niebles, J.C.:
  Action-agnostic human pose forecasting. 2019 IEEE Winter Conference on
  Applications of Computer Vision (WACV) pp. 1423--1432 (2019)

\bibitem{devlin-etal-2019-bert}
Devlin, J., Chang, M.W., Lee, K., Toutanova, K.: Bert: Pre-training of deep
  bidirectional transformers for language understanding. arXiv preprint
  arXiv:1810.04805  (2018)

\bibitem{dosovitskiy2020vit}
Dosovitskiy, A., Beyer, L., Kolesnikov, A., Weissenborn, D., Zhai, X.,
  Unterthiner, T., Dehghani, M., Minderer, M., Heigold, G., Gelly, S.,
  Uszkoreit, J., Houlsby, N.: An image is worth 16x16 words: Transformers for
  image recognition at scale. ICLR  (2021)

\bibitem{Giuliari2021TransformerNF}
Giuliari, F., Hasan, I., Cristani, M., Galasso, F.: Transformer networks for
  trajectory forecasting. 2020 25th International Conference on Pattern
  Recognition (ICPR) pp. 10335--10342 (2021)

\bibitem{Gong2011MultihypothesisMP}
Gong, H., Sim, J., Likhachev, M., Shi, J.: Multi-hypothesis motion planning for
  visual object tracking. 2011 International Conference on Computer Vision pp.
  619--626 (2011)

\bibitem{huangstgat}
Huang, Y., Bi, H., Li, Z., Mao, T., Wang, Z.: Stgat: Modeling spatial-temporal
  interactions for human trajectory prediction. In: Proceedings of the IEEE/CVF
  international conference on computer vision. pp. 6272--6281 (2019)

\bibitem{Koppula2013AnticipatingHA}
Koppula, H.S., Saxena, A.: Anticipating human activities for reactive robotic
  response. 2013 IEEE/RSJ International Conference on Intelligent Robots and
  Systems pp. 2071--2071 (2013)

\bibitem{Kosaraju2019SocialBiGATMT}
Kosaraju, V., Sadeghian, A., Mart{\'i}n-Mart{\'i}n, R., Reid, I.D.,
  Rezatofighi, S.H., Savarese, S.: Social-bigat: Multimodal trajectory
  forecasting using bicycle-gan and graph attention networks. In: NeurIPS
  (2019)

\bibitem{Levine2012ContinuousCC}
Levine, S., Wang, J.M., Haraux, A., Popovic, Z., Koltun, V.: Continuous
  character control with low-dimensional embeddings. ACM Transactions on
  Graphics (TOG)  \textbf{31},  1 -- 10 (2012)

\bibitem{Li2020EndtoendCP}
Li, L.L., Yang, B., Liang, M., Zeng, W., Ren, M., Segal, S., Urtasun, R.:
  End-to-end contextual perception and prediction with interaction transformer.
  2020 IEEE/RSJ International Conference on Intelligent Robots and Systems
  (IROS) pp. 5784--5791 (2020)

\bibitem{metro}
Lin, K., Wang, L., Liu, Z.: End-to-end human pose and mesh reconstruction with
  transformers. In: CVPR (2021)

\bibitem{Liu2021MultimodalMP}
Liu, Y., Zhang, J., Fang, L., Jiang, Q., Zhou, B.: Multimodal motion prediction
  with stacked transformers. 2021 IEEE/CVF Conference on Computer Vision and
  Pattern Recognition (CVPR) pp. 7573--7582 (2021)

\bibitem{AMASS:ICCV:2019}
Mahmood, N., Ghorbani, N., Troje, N.F., Pons-Moll, G., Black, M.J.: {AMASS}:
  Archive of motion capture as surface shapes. In: International Conference on
  Computer Vision. pp. 5442--5451 (Oct 2019)

\bibitem{Mao2020HistoryRI}
Mao, W., Liu, M., Salzmann, M.: History repeats itself: Human motion prediction
  via motion attention. In: ECCV (2020)

\bibitem{Mao2019LearningTD}
Mao, W., Liu, M., Salzmann, M., Li, H.: Learning trajectory dependencies for
  human motion prediction. 2019 IEEE/CVF International Conference on Computer
  Vision (ICCV) pp. 9488--9496 (2019)

\bibitem{Mao2021MultilevelMA}
Mao, W., Liu, M., Salzmann, M., Li, H.: Multi-level motion attention for human
  motion prediction. Int. J. Comput. Vis.  \textbf{129},  2513--2535 (2021)

\bibitem{vonMarcard20183dpw}
von Marcard, T., Henschel, R., Black, M., Rosenhahn, B., Pons-Moll, G.:
  Recovering accurate 3d human pose in the wild using imus and a moving camera.
  In: European Conference on Computer Vision (ECCV) (sep 2018)

\bibitem{MartinezGonzalez2021PoseT}
Mart'inez-Gonz'alez, A., Villamizar, M., Odobez, J.M.: Pose transformers
  (potr): Human motion prediction with non-autoregressive transformers. 2021
  IEEE/CVF International Conference on Computer Vision Workshops (ICCVW) pp.
  2276--2284 (2021)

\bibitem{singleshotmultiperson2018mupots}
Mehta, D., Sotnychenko, O., Mueller, F., Xu, W., Sridhar, S., Pons-Moll, G.,
  Theobalt, C.: Single-shot multi-person 3d pose estimation from monocular rgb.
  In: 2018 International Conference on 3D Vision (3DV). pp. 120--130. IEEE
  (2018)

\bibitem{Parsaeifarddecoupled}
Parsaeifard, B., Saadatnejad, S., Liu, Y., Mordan, T., Alahi, A.: Learning
  decoupled representations for human pose forecasting. In: Proceedings of the
  IEEE/CVF International Conference on Computer Vision. pp. 2294--2303 (2021)

\bibitem{lineardynamics}
Pavlovic, V., Rehg, J.M., MacCormick, J.: Learning switching linear models of
  human motion. Advances in neural information processing systems  \textbf{13}
  (2000)

\bibitem{Sadeghian2019SoPhieAA}
Sadeghian, A., Kosaraju, V., Sadeghian, A., Hirose, N., Savarese, S.: Sophie:
  An attentive gan for predicting paths compliant to social and physical
  constraints. 2019 IEEE/CVF Conference on Computer Vision and Pattern
  Recognition (CVPR) pp. 1349--1358 (2019)

\bibitem{Vaswani2017AttentionIA}
Vaswani, A., Shazeer, N., Parmar, N., Uszkoreit, J., Jones, L., Gomez, A.N.,
  Kaiser, {\L}., Polosukhin, I.: Attention is all you need. Advances in neural
  information processing systems  \textbf{30} (2017)

\bibitem{Wang2021SimpleBF}
Wang, C., Wang, Y., Huang, Z., Chen, Z.: Simple baseline for single human
  motion forecasting. 2021 IEEE/CVF International Conference on Computer Vision
  Workshops (ICCVW) pp. 2260--2265 (2021)

\bibitem{wang2021multi}
Wang, J., Xu, H., Narasimhan, M., Wang, X.: Multi-person 3d motion prediction
  with multi-range transformers. Advances in Neural Information Processing
  Systems  \textbf{34} (2021)

\bibitem{hierarchicalskeletaljoints}
Wu, D., Shao, L.: Leveraging hierarchical parametric networks for skeletal
  joints based action segmentation and recognition. In: Proceedings of the IEEE
  conference on computer vision and pattern recognition. pp. 724--731 (2014)

\bibitem{Yu2020SpatioTemporalGT}
Yu, C., Ma, X., Ren, J., Zhao, H., Yi, S.: Spatio-temporal graph transformer
  networks for pedestrian trajectory prediction. In: ECCV (2020)

\end{thebibliography}
}

\end{document}